\documentclass[runningheads]{llncs}
\usepackage[T1]{fontenc}
\usepackage{marvosym}
\usepackage{booktabs}
\usepackage{subscript}
\usepackage{amsmath}
\usepackage{pifont}
\usepackage{wasysym}
\usepackage{amssymb}
\usepackage{graphicx}
\begin{document}
\title{TripCast: Pre-training of Masked 2D Transformers for Trip Time Series Forecasting}
\titlerunning{TripCast}
%
\author{Yuhua Liao\textsuperscript{\Letter} \and
Zetian Wang \and Peng Wei \and Qiangqiang Nie \and Zhenhua Zhang}
\authorrunning{Y. Liao et al.}
%
\institute{Trip.com Group, Shanghai, China\\
\email{\{yh\_liao,zt\_wang,weip,qq\_nie,zhenhuazhang\}@trip.com}}
\maketitle              
\begin{abstract}
Deep learning and pre-trained models have shown great success in time series forecasting. However, in the tourism industry, time series data often exhibit a leading time property, presenting a 2D structure. This introduces unique challenges for forecasting in this sector. In this study, we propose a novel modelling paradigm, TripCast, which treats trip time series as 2D data and learns representations through masking and reconstruction processes. Pre-trained on large-scale real-world data, TripCast notably outperforms other state-of-the-art baselines in in-domain forecasting scenarios and demonstrates strong scalability and transferability in out-domain forecasting scenarios.

\end{abstract}
\section{Introduction}

\begin{figure}
\centering
\includegraphics[width=12cm]{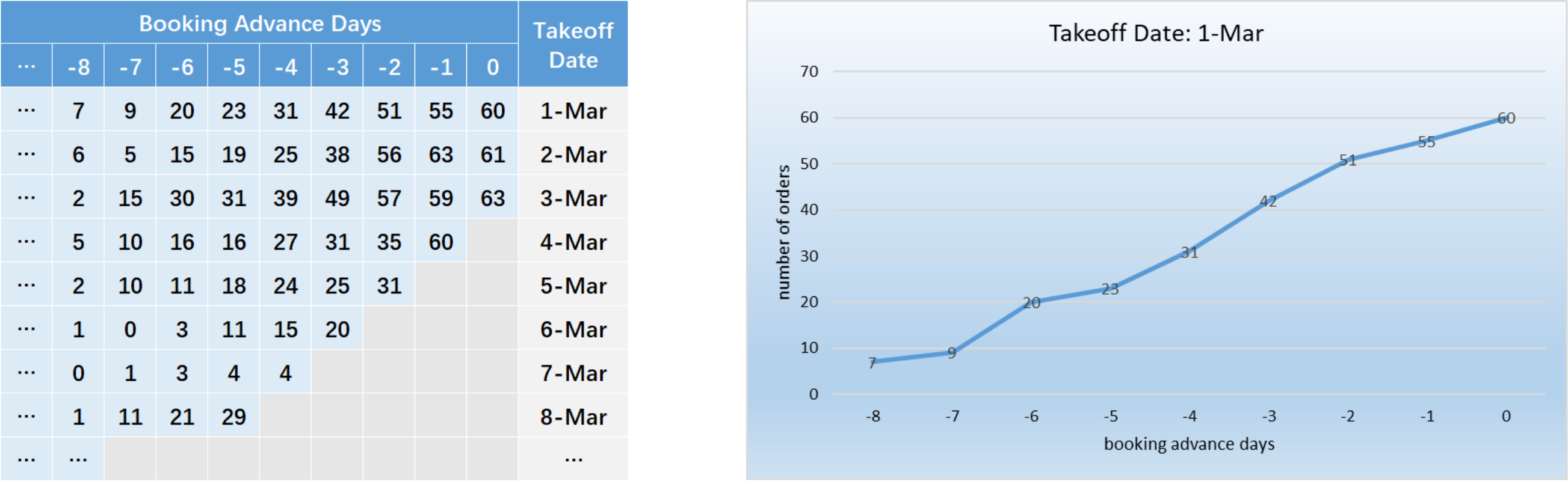}
\caption{An illustration of flight booking time series data (left). The vertical axis represents the flight takeoff date, and the horizontal axis represents the booking process. Within each takeoff date (right), the booking process is shown as a 1D time series and the entire data is shown as a 2D matrix. Across different takeoff dates, the unobserved booking process is shown as a triangle.} \label{fig1}
\end{figure}

Time series forecasting is widely used in various real-world fields, such as finance, speech analysis, action recognition, and traffic flow forecasting \cite{ma2023survey}. Accurate forecasts empower businesses to optimize decision-making, enhance operations, and improve overall efficiency \cite{bkaczek2023tspp}. In the tourism industry, time series forecasting plays a crucial role in revenue management \cite{hayes2021revenue}, demand planning \cite{kim2016forecasting}, and dynamic pricing \cite{pereira2016introduction}.

In the past decades, deep learning methods \cite{bai2018empirical,oreshkin2019n,zhou2021informer} have achieved significant success in time series forecasting \cite{ma2023survey}. These methods are flexible in modeling complex patterns and dependencies in time series, and have been widely used in various domains. However, training deep learning models from scratch requires a large amount of data and computational resources, which limits their usage in practice. In the tourism sector, new routes and flights are scheduled monthly without any historical data. Therefore, it is impractical to train a robust and accurate deep learning model for new routes or flights. More critically, in some domains, the application of deep time series models is hindered by the cold start problem due to the challenges or costs associated with data collection \cite{oreshkin2021meta}. Remarkably, large-scale pre-training has become a key element of training large neural networks in vision \cite{liu2024visual,radford2021learning} and text \cite{brown2020language,devlin2018bert} domain \cite{gruver2024large}. Large Language Models (LLMs) learn general representations from web-scale text data and both model size and data scale \cite{kaplan2020scaling} enhance corresponding zero-shot and in-context learning abilities. This inspires us to investigate the potential of pre-training time series models in the context of the tourism industry, especially given the limited research currently available in this field.

However, the time series data of the tourism industry inherently exhibits a dual-axis nature, as illustrated in Figure 1. The vertical axis represents the event time, such as the flight departure date, while the horizontal axis denotes the leading time prior to the event, such as the booking date. Existing forecasting paradigms typically address this problem from either the event time axis or the leading time axis. These dichotomous approaches result in two primary challenges: accuracy and efficiency.

Firstly, observations of time series in the tourism industry are typically influenced by both past event time points and leading time points. For instance, the booking rate of a flight on a specific departure date is influenced by the booking rate of the same flight on previous departure dates as well as the booking rate on previous leading times. Consequently, ignoring the complex dependencies and causality across different event times and leading times, existing models might fail to yield accurate forecasts. Secondly, building multiple models for different leading time steps or event time steps is inefficient and time-consuming. This fragmented approach necessitates significant computational resources and may lead to redundancy and suboptimal use of data. 

To address these challenges, we propose a novel modelling paradigm that treats trip time series as a whole 2D data, and learns local and global dependencies through masking and reconstruction training processes. Furthermore, to validate the transferability and scalability of TripCast as a zero-shot forecaster in the tourism industry, extensive experiments are conducted on zero-shot forecasting tasks in both in-domain and out-domain scenarios.

Our contributions are as follows:

* For the first time, we formulate the problem of trip time series forecasting and introduce a novel modelling paradigm that treats trip time series as 2D data to capture the intrinsic correlations and causality between different event times and leading times.

* To address the challenges of trip time series forecasting, we propose TripCast that learns local and global dependencies through masking and reconstruction processes.

* We perform comprehensive experiments based on large-scale datasets from an online travel agency. The results show that our method as a zero-shot forecaster, outperforms deep learning and pre-trained models in in-domain scenarios and achieves strong scalability and transferability in out-domain scenarios.

\section{Problem Statement}

\begin{figure}
\centering
\includegraphics[width=12cm]{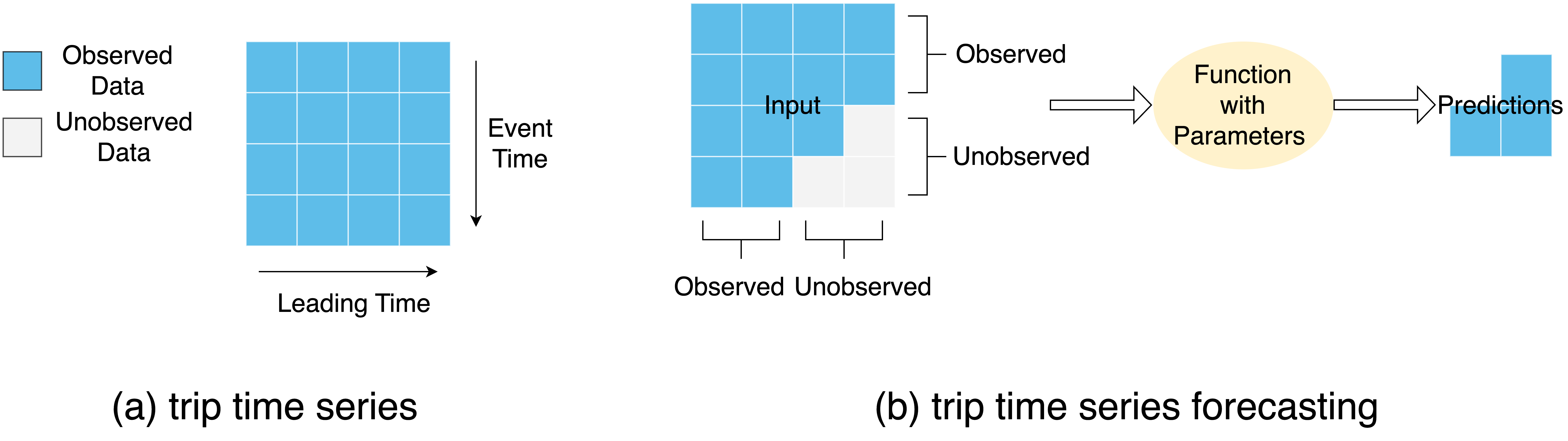}
\caption{Illustration of trip time series data (a) and trip time series forecasting problem (b).} \label{prob-f1}
\end{figure}

\subsection{Trip Time Series}
Let trip time series be denoted as sequential data with two axes, event time and leading time (Figure \ref{prob-f1}). The event time axis represents when a good or service is consumed, such as a flight takeoff date or a hotel room check-in date. The leading time axis represents the time before consumption, such as the booking date or search date. Formally, a trip time series $X$ is defined as a 2D matrix with dimensions $H \times C$, where $H$ is the length of the event time axis and $C$ is the length of the leading time axis. For simplicity, we ignore the covariates dimension in all definitions.

\subsection{Trip Time Series Forecasting}
Given a trip time series $X \in \mathbb{R}^{H \times C}$, $H_{obs}$ and $H_{pred}$ is the number of observed and predicted time steps along the event time axis. Correspondingly, $X$ has maximum $H_{pred}$ unobserved steps along the leading time axis and the number of unobserved leading steps is increasing with the advance of time. Our goal is to predict the unobserved leading time steps of future event time steps. Formally, the task can be defined as a parameterized function $\mathcal{F_{\theta}}$:

\begin{equation}
\mathcal{F}: X_{H_{obs}:, C_{obs}: } = \mathcal{F_{\theta}}(X_{: H_{obs}, :} \cup X_{H_{obs}:, :C_{obs}})
\end{equation}

where $C_{obs}$ are the observed leading time steps for each event time step. The problem is illustrated in Figure \ref{prob-f1}.

\subsection{In-domain and Out-domain Forecasting}

\begin{figure}
\centering
\includegraphics[width=8cm]{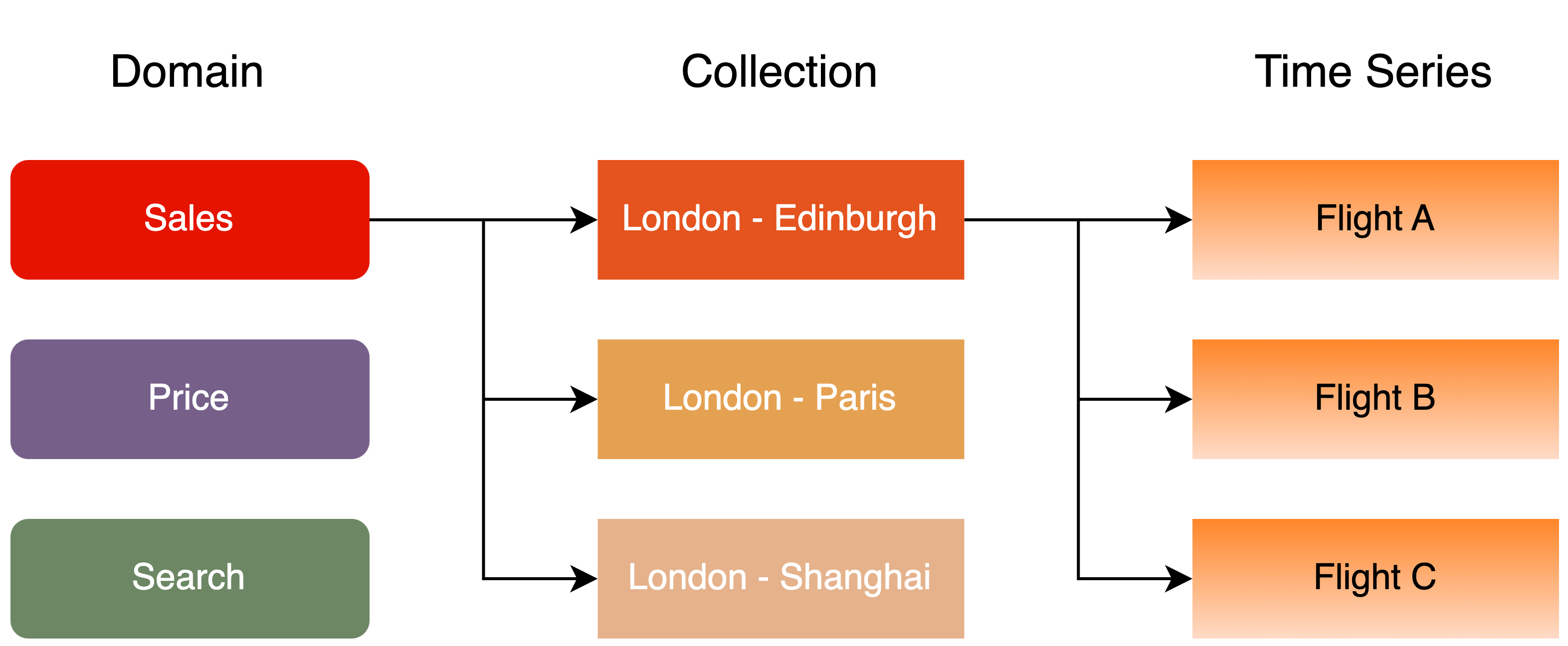}
\caption{Hierarchical granularities of trip time series. They can be categorized into three levels of granularities: domain, collection, and time series. Each domain contains multiple collections, and each collection contains multiple time series.} \label{fig1}
\end{figure}

Conceptually, temporal datasets can be categorized into three levels of granularities: domain, collection, and time series \cite{woo2023pushing} as shown in Figure \ref{fig1}. In-domain forecasting involves training and evaluating the model on the same dataset source. Conversely, out-domain forecasting entails training the model on multiple datasets and evaluating it on a dataset from a different domain. In this study, we focus on evaluating the effectiveness and scalability of TripCast in both in-domain and out-domain tasks.

\section{Related Work}
\subsection{Tourism Industry and Time Series Forecasting}

Time series forecasting is crucial in the tourism industry for revenue management, demand planning, and dynamic pricing. Existing forecasting methods can be classified into three categories: historical-data-based methods, advanced-data-based methods, and combined methods \cite{weatherford2003comparison}. Popular traditional methods in the tourism industry include ARIMA \cite{carmona2020sarima,do2020forecasting}, Exponential Smoothing \cite{yuksel2007integrated}, and Holt-Winters \cite{huang2023hotel}. With advancements in deep learning, some studies have explored leveraging deep learning models for tourism forecasting. In this work \cite{wang2022flight}, the authors trained forecasting models with temporal fusion transformer (TFT) \cite{lim2021temporal} for five different airports, and found that TFT outperforms traditional methods. 

\subsection{Pre-training Modelling for Time Series Analysis}

Inspired by advancements in pre-training across various fields, self-supervised learning has been adopted for time series forecasting. TS2Vec \cite{yue2022ts2vec} and CoST \cite{woo2022cost} learn representations
through contrastive learning. However, due to the limited scale of available datasets, they only consider in-domain scenarios, and their transferability is not well-studied. With the explosion of large language models (LLMs) \cite{ouyang2022training,touvron2023llama}, some studies explore to leverage LLMs for time series forecasting \cite{zhang2024large}. Time-LLM \cite{jin2023time} uses text data to reprogram time series modality into language modality. This approach \cite{zhou2023one} fine-tunes LLMs with time series datasets and achieves state-of-the-art performance on various forecasting scenarios. TEMPO \cite{cao2023tempo} introduces a prompt-based structure to enhance the distribution adaptation of LLMs for time series forecasting. Recently, foundation models pre-trained with time series data have been proposed \cite{das2023decoder,rasul2023lag,woo2024unified}.

\section{Methodology}

Within this section, we first outline the architecture of TripCast, which is well designed to accommodate the dual-axis properties of trip time series. We then describe the training strategies for both pre-training and downstream tasks.
\subsection{Model Structure}

\begin{figure}
\centering
\includegraphics[width=12cm]{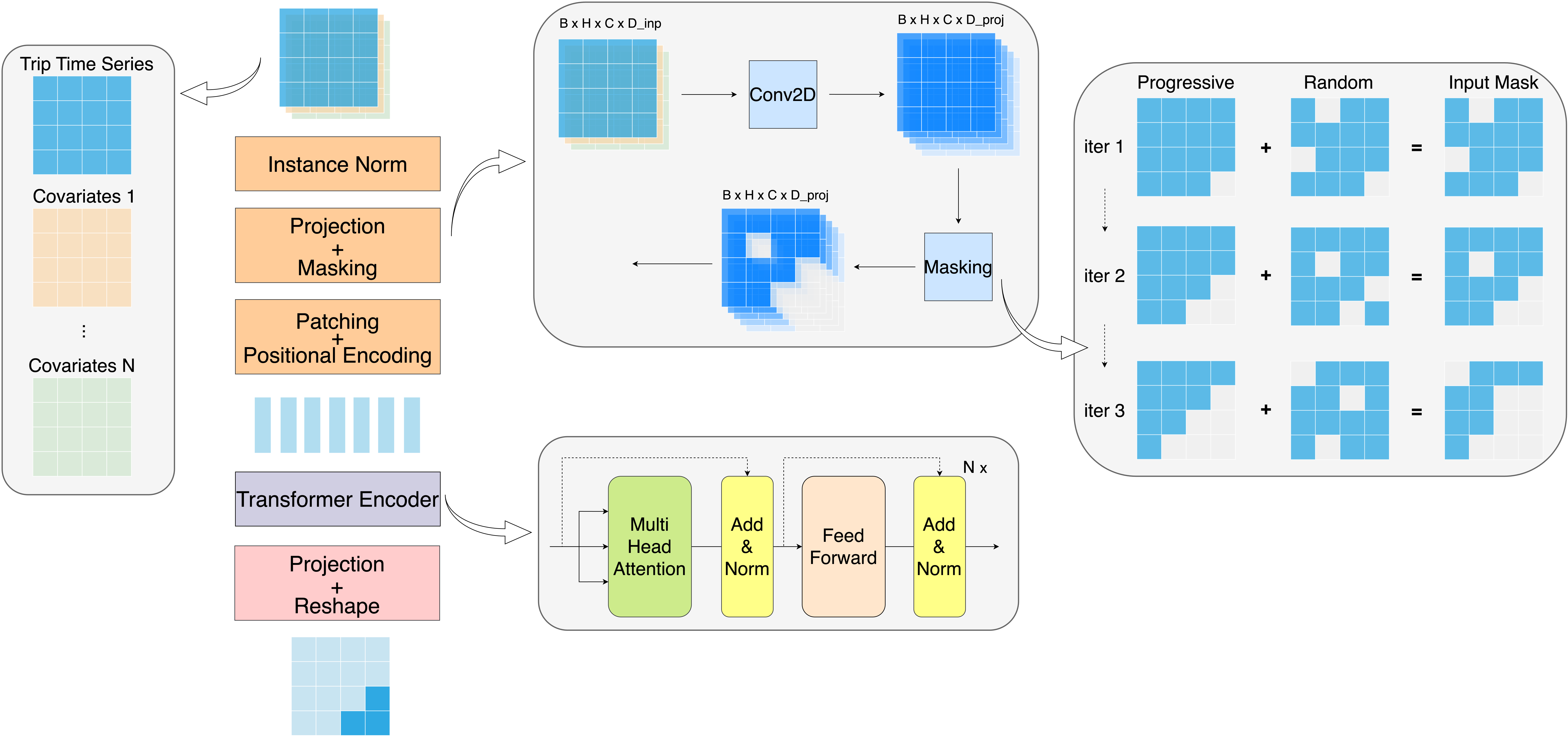}
\caption{The architecture of proposed TripCast model. Trip time series and covariates are stacked along the event time and leading time axes. The input data is normalized and projected to higher dimension. Then, token-level masking is applied to the projected input data. The masked input data is patched and fed into multiple transformer layers to learn predictive representations. Finally, output of the transformer layers is projected and reconstructed to estimate the unobserved values of future time steps.} \label{model}
\end{figure}

\noindent\textbf{Input Projection and Masking.} Unlike image modeling \cite{dosovitskiy2020image}, we cannot directly apply patch masking to trip time series because observed and unobserved values might be mixed within the same patch. To tokenize the unobserved and missing values, we are inspired by TS2Vec \cite{yue2022ts2vec} and project the input $x_{h, c}$ to a higher dimension latent vector $z_{h, c}$ and apply token-level mask to the input data. Notably, we mask the latent vectors rather than the raw input data because the value range of the raw input data is dynamic, making it impractical to use a fixed mask value. In this way, observed and unobserved tokens are separated in the latent representations space. Furthermore, we adopt two masking strategies during pre-training stage:

* Random masking: This strategy simulates missing data by masking a predetermined proportion of tokens from the projected data at random (Figure \ref{model}). It enhances the robustness of TripCast models and ensures stable performance in real-world applications.

$$
Mask^{random}_{h, c} \sim Bernoulli(p), \hspace{1cm} Mask^{random}_{h, c} \in \{0, 1\}
$$

* Progressive masking: In trip time series, unobserved values typically appear in a triangular form, and with the progress of time, unobserved values along the diagonal are gradually revealed. To inject this prior knowledge into training stage and help the model learn causality, we mask triangular regions of the input data in a progressive manner which is shown in Figure \ref{model}.

During inference stage, we only mask the unobserved tokens and feed the masked input data into the model to predict these values.\\
\textbf{Patching and Positional Encoding.} As demonstrated by PatchTST \cite{nie2022time} and Vision Transformer \cite{dosovitskiy2020image}, patching is an effective way to capture local patterns. In TripCast, we segment the input data into non-overlapping patches and apply a linear projection to each patch. This process reduces input data redundancy and extracts local semantic information. To capture the order of the input sequence, we use sinusoidal positional encoding to encode the positional information of the input data.

\begin{equation}
z^{pacth} = PatchEmbed(z) + SinusoidalPositionalEncoding(z)
\end{equation}\\
\textbf{Transformer Encoder.} After patching and positional encoding, we use standard transformer encoder to map the input tokens to latent representations. Each of these layers is composed of a multi-head self-attention mechanism and subsequently a feed-forward neural network. 

\begin{gather}
z_1^{enc} = SelfAttention(z^{patch}) \\
z_2^{enc} = LayerNorm(z_1^{enc} + z^{patch}) \\
z_3^{enc} = FeedForward(z_2^{enc}) \\
z^{enc} = LayerNorm(z_3^{enc} + z_2^{enc})
\end{gather}
\textbf{Reconstruction.} Given the latent representations of the transformer encoder, we project the latent vector to $P \times P \times N$, where $P$ is the size of patch and $N$ is number of predicted series. In this work, we focus on univariate scenario, so $N$ is 1. Then, we reshape the projected latent vectors to $(B, H, C, N)$ as the output of model.\\
\textbf{Instance Normalization.} To mitigate the distribution drift between training and test data, we apply reversible instance normalization \cite{kim2021reversible} in TripCast models. This normalization module scales the input data by the mean and variance, then reverses the scaling for the output predictions. Although our input data is 2D, the mean and variance are calculated in the same manner as in typical instance normalization.

\subsection{Pre-training and Downstream Tasks}

We split each dataset into pre-train and train-test partitions in a roughly 90/10 split. To prevent data leakage, we ensure that all return routes and flights are either in the pre-train or train-test set. For train-test sets, we choose the data from 2019-06-01 to 2019-08-31 as validation set and the data from 2019-09-01 to 2019-12-31 as test set on all datasets. All TripCast models are trained on pre-train datasets and evaluated on train-test datasets. Our aim is to demonstrate the potential of TripCast as a zero-shot forecaster in the tourism industry.\\
\textbf{Pre-training.} In this work, we focus on supervised pre-training since our main goal is to demonstrate the effectiveness and transferability of this novel modelling paradigm. In all pre-training tasks, we set $H$ to 60, $C$ to 40 and the maximum $H_{pred}$ of progressive masking to 15. Furthermore, we use mean absolute error (MAE) as the loss function to train the model during the pre-training stage.\\
\textbf{Downstream Tasks.} After pre-training, we evaluate TripCast on two downstream tasks: in-domain forecasting and out-domain forecasting. In in-domain forecasting, we pre-train the model within each domain and assess its performance within the same domain. In out-domain forecasting, we pre-train a unified model on all domains, then evaluate its performance on each domain.
\section{Experiments}

In this work, we collect five extensive, real-world datasets from an online travel agency (OTA) to evaluate the performance of TripCast. These collections encompass flight sales data, flight booking price data, and user search data. First, we pre-train TripCast models of small and base sizes on each dataset and evaluate their performance in in-domain forecasting scenarios. Next, we compare our method with deep learning and pre-trained time series models. Then, for investigating the transferability as well as scalability of TripCast models, we pre-train TripCast model of large size on four datasets except UserSearch, and evaluate its performance on out-domain forecasting tasks. Finally, we conduct extensive ablation studies and examine the impact of various components and masking strategies on the performance of TripCast.
\subsection{Datasets}
All datasets are preprocessed into univariate time series with date features. Below are the details of the datasets:

\begin{table}[]
\begin{tabular}{ccccccc}
\hline
\textbf{Dataset} & \textbf{Period}      & \multicolumn{2}{c}{\textbf{Total n\_series}} & \multicolumn{2}{c}{\textbf{Total n\_obs}} & \textbf{Frequency} \\ \hline
                 &                      & Pre-train            & Train-test            & Pre-train            & Train-test         &                    \\
FlightSales      & 2018-01$\sim$2019-12 & 3,947                 & 489                   & 110,997,640          & 13,712,640         & Day                \\
RouteSales       & 2018-01$\sim$2019-12 & 2,572                 & 286                   & 68,789,440           & 7,626,080          & Day                \\
FlightPrice      & 2017-08$\sim$2019-12 & 5,395                 & 595                   & 173,911,800          & 19,237,680         & Day                \\
RoutePrice       & 2017-01$\sim$2019-12 & 3,996                 & 445                   & 159,749,040          & 17,685,080         & Day                \\
UserSearch       & 2017-04$\sim$2019-12 & 3,298                 & 367                   & 124,884,320          & 13,690,880         & Day                \\ \hline
\end{tabular}
\caption{Key details of datasets.}
\label{tab:my-table}
\end{table}

* FlightSales: This dataset contains the daily sales rate of seats which is the ratio of the number of seats sold to the capacity of flights. All time series in this dataset are aggregated by flight.

* RouteSales: This dataset is similar to FlightSales, but the time series are aggregated by route.

* FlightPrice: This dataset contains the accumulative average order price of flights. All time series in this dataset are aggregated by flight.

* RoutePrice: This dataset is similar to FlightPrice, but the time series are aggregated by route.

* UserSearch: This dataset contains the accumulative user search count of routes.

\subsection{Training}
We pre-train the models in three different sizes ranging from small to large, with detailed hyperparameters shown in table \ref{tab:modelsize}. The minimum model has less than 1 million parameters while the large model has nearly 20 million parameters. All models are trained with a batch size of 256 and 50000 iterations. We use Adam \cite{kingma2014adam} with an initial learning rate of 3e-4, and cosine learning rate decay. The training is conducted using NVIDIA V100 GPUs with mixed precision training.

\begin{table}[]
\centering
\begin{tabular}{@{}ccclcc@{}}
\toprule
\textbf{Model}  & \textbf{Layers} & \textbf{Dimension} & \textbf{Heads} & \textbf{Params} & \textbf{Iters} \\ \midrule
TripCast\textsubscript{small} & 4               & 128                & 4              & 928k            & 50000          \\
TripCast\textsubscript{base}  & 4               & 256                & 8              & 3.4m           & 50000          \\
TripCast\textsubscript{large} & 6               & 512                & 8             & 19.4m           & 50000          \\ \bottomrule
\end{tabular}
\caption{Details of the hyperparameters of TripCast models in different sizes.}
\label{tab:modelsize}
\end{table}

\subsection{Evaluation Metrics}

As evaluation criteria, in this study, we employ mean absolute error (MAE) and weighted absolute percentage error (WAPE).

\begin{equation}
MAE = \frac{1}{n} \sum_{i=1}^{n} |y_i - \hat{y}_i|; \quad
\mathrm{WAPE}=\frac{\sum_{i}\left|y_{i}-\hat{y}_{i}\right|}{\sum_{i}\left|y_{i}\right|}
\end{equation}

\subsection{Baselines}

For deep learning methods, we compare TripCast with linear family \cite{zeng2023transformers}, iTransformer \cite{liu2023itransformer}, and PatchTST \cite{nie2022time}. For pre-trained models, we compare TripCast with GPT4TS \cite{zhou2023one}. The details of the baselines are as follows:

\begin{table}[]
\centering
\begin{tabular}{@{}ccc@{}}
\toprule
Baseline      & Hyperparameters                                                                        & Values                                                                               \\ \midrule
LinearFamily & model type                                                                             & \{Linear, NLinear, DLinear\}                                                         \\ \midrule
PatchTST     & \begin{tabular}[c]{@{}c@{}}d\_model\\ num\_layers\end{tabular}                         & \begin{tabular}[c]{@{}c@{}}\{128, 256\}\\ \{2, 3, 4\}\end{tabular}                   \\ \midrule
iTransformer & \begin{tabular}[c]{@{}c@{}}d\_model\\ num\_layers\\ use\_norm\end{tabular}             & \begin{tabular}[c]{@{}c@{}}\{128, 256\}\\ \{2, 3, 4\}\\ \{true, false\}\end{tabular} \\ \midrule
GPT4TS       & \begin{tabular}[c]{@{}c@{}}block\_size\\ n\_head\\ d\_model\\ num\_layers\end{tabular} & \begin{tabular}[c]{@{}c@{}}\{1024\}\\ \{12\}\\ \{768\}\\ \{6\}\end{tabular}          \\ \bottomrule
\end{tabular}
\caption{Hyperparameter search range for baselines.}
\label{tab:my-table}
\end{table}
Constrained by the fact that all baselines are single-axis time series models, we simplify the forecasting task to predicting the value of the last leading step for convenience. The look-back period and prediction horizon of baselines are set to 45 and 15, which are consistent with TripCast models. This setting ensures that the performance of both TripCast and baselines is evaluated at the same time points. Additionally, with a batch size of 256, training of deep learning baselines is conducted over 10,000 iterations. Based on validation loss, early stopping is implemented, with the loss being summarized and reported at intervals of 100 iterations. The optimal checkpoint is chosen according to the validation loss. For pre-trained models, we use the same training hyperparameters as TripCast models. In summary, deep learning models are trained from scratch on train-test datasets, while pre-trained models are trained on pre-train datasets and follow zero-shot evaluation on train-test datasets.
\section{Results}

\subsection{In-domain Forecasting}

The performance of TripCast models and baselines in in-domain scenarios is illustrated in Table \ref{tab:indomain}. We find that both TripCast\textsubscript{small} and TripCast\textsubscript{base} outperform all baselines across all datasets. Among deep learning methods, PatchTST outperforms other methods in three out of five datasets indicating that patching and transformer-based models effectively capture trip time series patterns. GPT4TS, as a LLM-based model outperforms deep learning methods in three out of five datasets. We speculate that the strong transferability of GPT2 and the extensive pre-training data contribute to its superior performance. This also highlights the potential of pre-trained models in trip time series forecasting.

\begin{table}[]
\centering
\begin{tabular}{@{}ccccccccccc@{}}
\toprule
                & \multicolumn{2}{c}{FlightSales} & \multicolumn{2}{c}{RouteSales}  & \multicolumn{2}{c}{FlightPrice} & \multicolumn{2}{c}{RoutePrice}  & \multicolumn{2}{c}{RouteSearch} \\ \midrule
                & MAE            & WAPE           & MAE            & WAPE           & MAE            & WAPE           & MAE            & WAPE           & MAE            & WAPE           \\ \midrule
Linear          & 0.064          & 0.193          & 0.048          & 0.153          & 116.8          & 0.151          & 167.4          & 0.192          & 94.3           & 0.127          \\
NLinear         & 0.063          & 0.193          & 0.048          & 0.153          & 115.7          & 0.149          & 169.1          & 0.195          & 95.3           & 0.129          \\
DLinear         & 0.064          & 0.193          & 0.048          & 0.152          & 113.1          & 0.146          & 166.7          & 0.191          & 92.2           & 0.124          \\
PatchTST        & 0.064          & 0.193          & 0.048          & 0.155          & 109.7          & 0.142          & 162.3          & 0.186          & 88.8           & 0.119          \\
iTransformer    & 0.064          & 0.193          & 0.048          & 0.152          & 110.8          & 0.143          & 163.1          & 0.187          & 90.3           & 0.121          \\ \midrule
GPT4TS          & 0.063          & 0.193          & 0.047          & 0.152          & 110.0          & 0.142          & 161.2          & 0.185          & 79.9           & 0.108          \\ \midrule
TripCast\textsubscript{small} & 0.052          & 0.159          & \textbf{0.038} & 0.122          & 94.7           & 0.122          & 106.6          & 0.127          & 47.2           & 0.064          \\
TripCast\textsubscript{base}  & \textbf{0.050} & \textbf{0.153} & \textbf{0.038} & \textbf{0.121} & \textbf{91.4}  & \textbf{0.118} & \textbf{103.7} & \textbf{0.124} & \textbf{44.5}  & \textbf{0.061} \\ \bottomrule
\end{tabular}
\caption{Test set results for deep learning and pre-trained baseline methods. Optimal results are highlighted in bold.}
\label{tab:indomain}
\end{table}

\subsection{Towards Foundation Model (Out-domain Forecasting)}

\begin{figure}
\centering
\includegraphics[width=9cm]{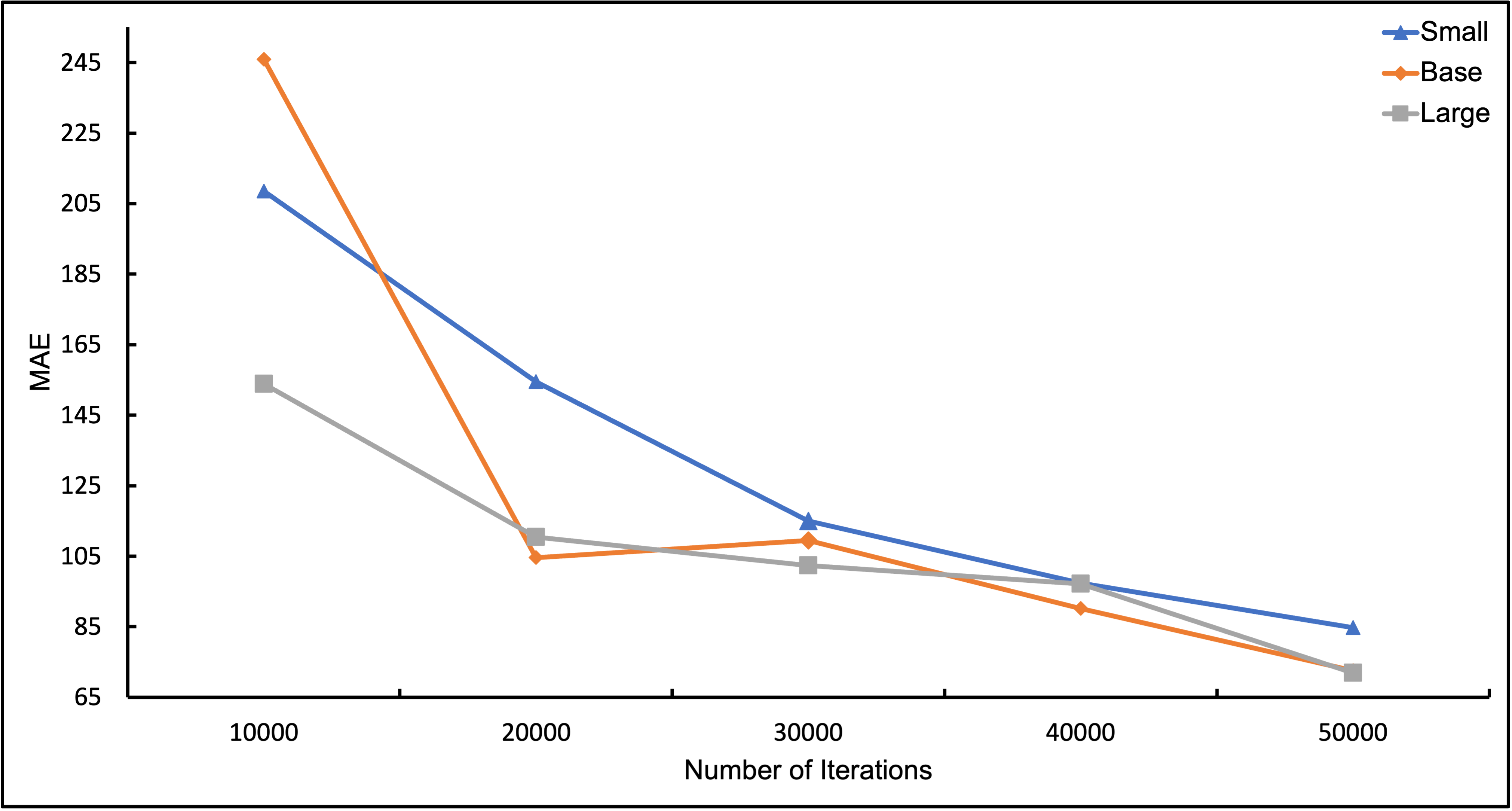}
\caption{Accuracy versus the number of iterations during pre-training for different model sizes.} \label{out-domain}
\end{figure}

The ultimate goal of our research is to develop a foundation model for trip time series forecasting. Experimentally, we investigate the effectiveness of our model in out-domain forecasting. We pre-train model of different sizes (Figure \ref{out-domain}) on all datasets except UserSearch and evaluate their performance on UserSearch dataset. Our findings indicate that TripCast models perform well on the UserSearch dataset. The accuracy of TripCast\textsubscript{small} is close to PatchTST, while TripCast\textsubscript{base} and TripCast\textsubscript{large} outperforms GPT4TS although it is pre-trained on target domain. Furthermore, we observe that TripCast models' performance scales well with the number of training iterations. This suggests that our method is a promising candidate for a foundational model in trip time series forecasting.

\subsection{Ablation Study}

\subsubsection{Masking Strategy.}
We conducted ablation studies on masking strategy, with a focus on progressive masking, as the robustness of the model is not our primary concern in this work. Table \ref{tab:ablation-mask} shows that dynamic progressive masking helps models learn causality and achieve better performance.

\subsubsection{Positional Encoding.}
Attention mechanism is permutation invariant, so transformer models rely on positional encoding to capture the order of the input sequence. We compared the performance of TripCast\textsubscript{base} with learned positional encoding, fixed positional encoding, and no positional encoding. Our findings, summarized in Table \ref{tab:ablation-pos}, indicate that fixed positional encoding yields better performance than learned positional encoding.

\begin{table}[]
\centering
\begin{tabular}{@{}ccccccc@{}}
\toprule
                     & \multicolumn{2}{c}{FlightSales} & \multicolumn{2}{c}{FlightPrice} & \multicolumn{2}{c}{RouteSearch} \\ \midrule
                     & MAE            & WAPE           & MAE            & WAPE           & MAE            & WAPE           \\ \midrule
TripCast\textsubscript{base}       & \textbf{0.050} & \textbf{0.153} & \textbf{91.4}  & \textbf{0.118} & \textbf{44.5}  & \textbf{0.061} \\
w/o Progressive Mask & 0.051          & \textbf{0.153} & 92.1           & 0.119          & 45.3           & 0.062          \\ \bottomrule
\end{tabular}
\caption{Ablation study of the masking strategy.}
\label{tab:ablation-mask}
\end{table}

\begin{table}[]
\centering
\begin{tabular}{@{}ccccccccc@{}}
\toprule
\multicolumn{3}{l}{}  & \multicolumn{2}{c}{FlightSales} & \multicolumn{2}{c}{FlightPrice} & \multicolumn{2}{c}{RouteSearch} \\ \midrule
Date/Time & SPE & LPE & MAE            & WAPE           & MAE            & WAPE           & MAE            & WAPE           \\ \midrule
\checkmark         &     &     & 0.062          & 0.186          & 99.2           & 0.127          & 79.5           & 0.107          \\
\checkmark         & \checkmark   &     & \textbf{0.050} & \textbf{0.153} & 91.4           & 0.118          & \textbf{44.5}  & \textbf{0.061} \\
\checkmark         &     & \checkmark   & 0.052          & 0.157          & \textbf{90.1}  & \textbf{0.116} & 49.7           & 0.067          \\ \bottomrule
\end{tabular}
\caption{Ablation study of the positional encoding.}
\label{tab:ablation-pos}
\end{table}

\section{Conclusion}

In this study, the trip time series forecasting problem is formulated and we proposed a novel modelling paradigm to tackle its challenges. We pre-train transformer-based models on five large-scale real-world datasets and subsequently evaluate their performance in in-domain forecasting. Our findings demonstrate the effectiveness of our approach against other deep learning and pre-trained models. Additionally, we show that our method scales well with model size and training iterations for out-of-domain forecasting. Our work opens new possibilities for time series forecasting in tourism, and we hope that it will inspire further research in this area.

%
%
%
\bibliographystyle{splncs04}
\bibliography{6235}

\end{document}